\documentclass[10pt,twocolumn,letterpaper]{article}

\usepackage{cvpr}              % To produce the CAMERA-READY version
\usepackage{xcolor}
\usepackage{array}
\usepackage{amsmath}
\usepackage{stfloats}
\usepackage{graphicx}
\usepackage{booktabs}
\usepackage{caption}
\usepackage{multirow}
\usepackage{makecell}
\usepackage{colortbl}
\usepackage[table]{xcolor}
\usepackage{graphicx}
\usepackage{subcaption}
\usepackage{booktabs} % 表格美化
\usepackage{xcolor}   % 颜色
\usepackage[most]{tcolorbox} % 用于制作好看的文本框
\definecolor{cgreen}{RGB}{0,150,0} % 定义绿色用于标记正确答案
\definecolor{cred}{RGB}{200,0,0}   % 定义红色用于标记错误答案

\definecolor{cvprblue}{rgb}{0.21,0.49,0.74}
\usepackage[pagebackref,breaklinks,colorlinks,allcolors=cvprblue]{hyperref}
\def\paperID{} % *** Enter the Paper ID here
\def\confName{CVPR}
\def\confYear{2026}

\title{HAWK: Head Importance-Aware Visual Token Pruning in Multimodal Models}

\author{Qihui Zhu\quad Tao Zhang\quad Yuchen Wang \quad Zijian Wen \quad Mengjie Zhang \quad Shuangwu Chen\thanks{Corresponding author} \\
\quad Xiaobin Tan\quad Jian Yang\\
University of Science and Technology of China
\and
Yang Liu\\
ChangXin Memory Technologies, Inc
\and
Zhenhua Dong\quad Xianzhi Yu\quad Yinfei Pan\\
Huawei Noah's Ark Lab
}

\begin{document}
\maketitle
\begin{abstract}
In multimodal large language models (MLLMs), the surge of visual tokens  significantly increases the inference time and computational overhead, making them impractical for real-time or resource-constrained applications.
Visual token pruning is a promising strategy for reducing the cost of MLLM inference by removing redundant visual tokens.
Existing researches usually assume that all attention heads contribute equally to the visual interpretation.
However, our study reveals that different heads may capture distinct visual semantics and inherently  play distinct roles  in visual processing.
In light of this observation, we propose HAWK, a \textbf{h}ead importance-\textbf{aw}are visual to\textbf{k}en pruning method that perceives the varying importance of attention heads in visual tasks to maximize the retention of crucial tokens.
By leveraging head importance weights and text-guided attention to assess visual token significance, HAWK effectively retains task-relevant visual tokens while removing redundant ones.
The proposed HAWK  is entirely training-free and can be seamlessly applied to various MLLMs.
Extensive experiments on multiple mainstream vision-language benchmarks demonstrate that HAWK achieves state-of-the-art accuracy. When applied to Qwen2.5-VL, HAWK retains 96.0\% of the original accuracy after pruning 80.2\% of the visual tokens. Additionally, it reduces end-to-end latency to 74.4\% of the original and further decreases GPU memory usage across the tested models.The code is available at https://github.com/peppery77/HAWK.git.
\end{abstract}    
\section{Introduction}
\label{sec:intro}
\begin{figure}[t]
    \centering
    \includegraphics[width=0.9\linewidth]{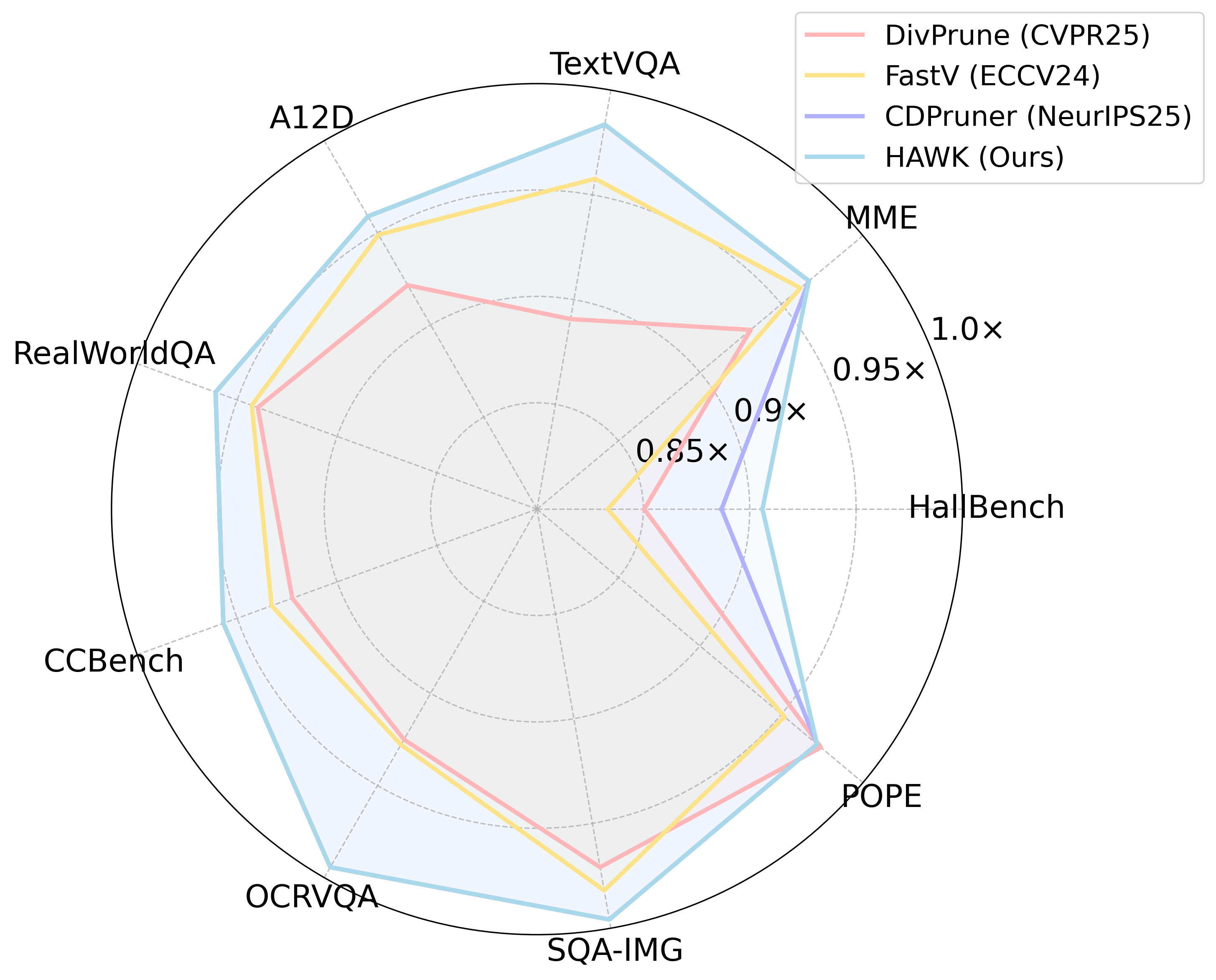}
    \caption{Comparison of visual token pruning methods on multiple benchmarks for Qwen2.5-VL-7B, where the proposed approach consistently outperforms all baselines.}
    \label{fig:overall-performance}
\end{figure}

\begin{figure*}[t]
    \centering
    \includegraphics[width=1\linewidth]{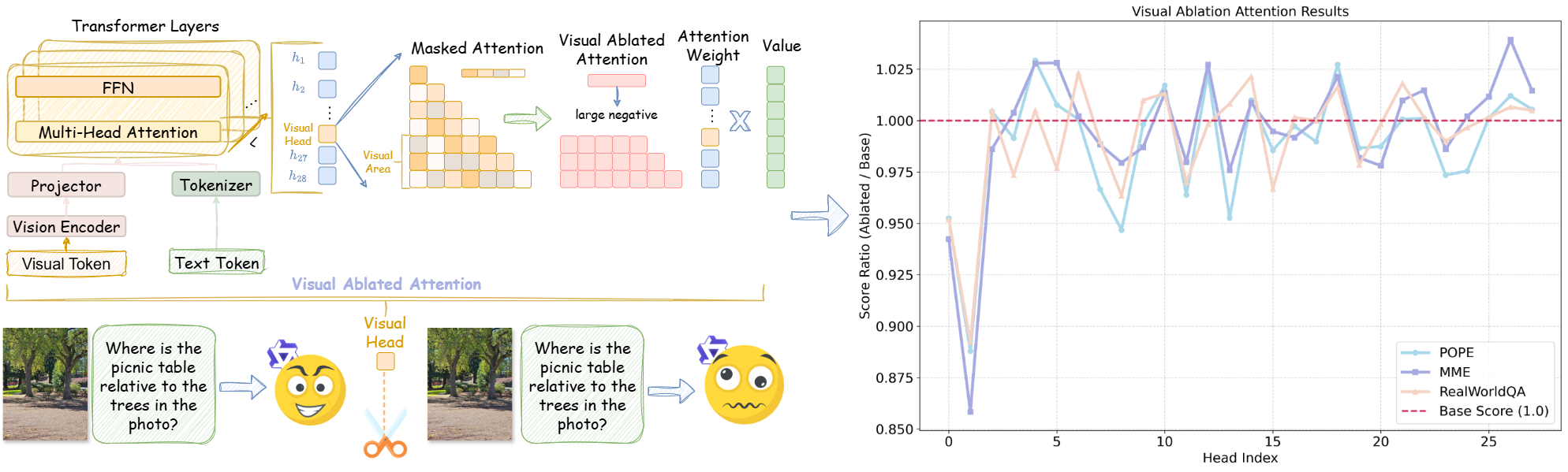}
    \caption{\textbf{Left.} The ablation process of the visual attention head. After ablating the attention heads as described above, the model’s visual comprehension ability noticeably declines. \textbf{Right.} Ablation results of visual attention heads. Different heads exhibit varying impacts on visual tasks, and their importance shows consistent trends across multiple datasets.}
    \label{fig:ablation-method}
\end{figure*}

Recent advances in Large Language Models (LLMs)\cite{llama2, vicuna, gpt4, qwen} have driven rapid progress in Multimodal Large Language Models (MLLMs)\cite{visual-tuning, visual-tuning2, llava, qwen2vl, Internvl}, extending their powerful reasoning abilities to visual modalities such as images and videos. To fully harness the strengths of LLMs, MLLMs typically encode visual inputs into a language-like representation known as visual tokens. 
The number of visual tokens often reaches the hundreds, far exceeding that of textual tokens,  and would grow even larger with higher resolutions \cite{llavanext, llava-uhd} or longer videos\cite{video-llama, video-tuning}. 
As the computational complexity of attention-based models grows quadratically with the token length\cite{transformer}, the surge of visual tokens in MLLMs significantly increases the inference time and computational overhead, making them impractical for real-time or resource-constrained applications\cite{minicpm, gemma}.

Visual token pruning is a promising strategy for reducing the cost of MLLM inference by removing redundant visual tokens. Existing researches can be broadly grouped into three categories:  similarity-based,  fine-tuning-based and attention-based methods.  
Similarity-based methods \cite{cdpruner, divprune, similarity, similarity2} remove redundant regions by measuring feature similarity among visual tokens.
However, they operate in a context-agnostic manner disregarding user instructions, and thereby fail to adapt to the wide variety of queries or tasks encountered in real-world scenarios.
Fine-tuning-based methods \cite{fine-tuning, fine-tuning2} incorporate  pruning as a learnable module that is jointly optimized with the model parameters.
Despite their effectiveness, they rely on end-to-end training, suffering from high computational overhead and reduced generalization across different  tasks.
Attention-based methods \cite{FastV, pyramiddrop, prumerge} estimate token importance from early-layer visual attention scores.
These methods usually assume that all attention heads contribute equally to the visual interpretation and  simply perform arithmetic averaging over head weights.

However, in fact, as different heads may capture distinct visual semantics, they
should inherently  exhibit distinct importance in visual processing, which is not considered by existing researches.
As shown in Fig.~\ref{fig:ablation-method}, to investigate the roles of each individual attention head in visual understanding, we systematically disabled each head’s access to visual tokens and measured the impact on model performance across multiple benchmark datasets. 
Masking different attention heads leads to markedly different variations in model performance, and this variation trend is consistent across different datasets.
The results reveal that  each attention head plays a unique and distinct role in visual processing within MLLMs. Treating different attention heads equally may  lead to the retention of redundant tokens while inadvertently pruning semantically valuable ones.

This observation motivates us to integrate head importance into the visual token pruning. Accordingly, we propose HAWK, a head importance-aware visual token pruning method for accelerating MLLM inference. 
%It performs pruning based on text-guided visual token importance while incorporating attention head importance weights. 
% HAWK 工作原理
HAWK leverages an attention re-weighting mechanism that explicitly accounts for head importance in visual token pruning. HAWK initially leverages a text-guided attention mechanism to adaptively estimate the semantic relevance between visual tokens and the textual input. Then, it performs a one-time offline analysis to estimate the intrinsic contribution of each attention head to visual understanding. During inference, these precomputed head importance weights are applied to weight the visual token scores, thereby enabling more accurate estimation of their importance for guiding pruning.
Consistent with all advanced pruning approaches, HAWK is entirely training-free and requires only a single offline computation. It eliminates the need for costly calibration or fine-tuning, ensuring broad applicability across diverse MLLM architectures such as LLaVA-NeXT and Qwen2.5-VL. As shown in Fig.~\ref{fig:overall-performance}, HAWK consistently outperforms existing pruning baselines across multiple benchmarks. Extensive experiments on multiple vision-language benchmarks demonstrate that HAWK achieves state-of-the-art (SOTA) performance. When applied to Qwen2.5-VL, HAWK preserves 96.0\% of the original performance while pruning 80.2\% of visual tokens. Moreover, it reduces end-to-end inference latency by 26\%  compared to the original.
In summary, the main contributions of this work can be summarized as follows:
\begin{itemize}
    \item We find that different attention heads in MLLMs play distinct roles in visual processing, and we further quantify the importance of each head.
    \item We propose HAWK, a head importance–aware visual token pruning method that is entirely training-free.
    \item We conduct extensive experiments on multiple vision-language benchmarks, showing that HAWK consistently achieves SOTA performance across various pruning ratios. % E2E 时延 精度减少
\end{itemize}

\section{Related work}
\label{sec:formatting}
\subsection{Multimodal Large Language Models}

The remarkable success of large language models (LLMs)\cite{llm-app, llm-app2, qwen, qwen2vl, Internvl} has motivated extensive research to extend their reasoning capabilities to other modalities, leading to the emergence of multimodal large language models (MLLMs)\cite{visual-tuning, visual-tuning2, llava, qwen2vl, Internvl}. By integrating visual encoders with pretrained language backbones, MLLMs enable unified cross-modal understanding and reasoning, supporting a wide range of tasks such as visual question answering, image captioning, and video comprehension. In these models, visual inputs are typically converted into sequences of tokens and processed alongside text tokens within the LLM, allowing for joint reasoning across modalities. However, this design introduces a major computational bottleneck. Visual information is inherently dense and spatially redundant, leading to significantly more tokens than text. For instance, LLaVA-1.5\cite{visual-tuning2} converts a 336×336 image into 576 tokens, while its high-resolution successor LLaVA-NeXT\cite{llavanext} supports AnyRes technology and Qwen2.5VL\cite{qwen} supports NativeRes technology. The problem is further amplified in video understanding, where models such as LongVA\cite{longva} generate over 200K tokens from 2,000 frames—comparable to the context length of advanced text-only LLMs. This excessive token load drastically increases computational and memory demands, limiting scalability and practical deployment. Consequently, visual token pruning has emerged as an essential strategy for improving inference efficiency in MLLMs. By removing redundant or less informative tokens while preserving task-relevant content, these methods aim to reduce the computational burden without compromising accuracy, representing a key direction toward scalable and efficient multimodal reasoning.

\subsection{Visual Token Pruning}

Recent studies have proposed numerous visual token pruning methods to accelerate multimodal large language model inference by removing redundant visual tokens\cite{FastV, divprune, cdpruner, similarity, similarity2, fine-tuning, fine-tuning2}. These approaches can be roughly divided into four groups. First, importance-based methods\cite{FastV, Fitprune, sparsevlm} rank tokens by attention-derived relevance and prune those with low importance. Although simple and training-free, they often retain clusters of redundant tokens in salient regions. Second, similarity-based methods, such as DivPrune\cite{divprune}, maximize feature diversity to ensure broader spatial coverage, but their instruction-agnostic design may discard task-critical information. Third, hybrid approaches combine importance and diversity to balance relevance and coverage. CDPruner\cite{cdpruner} exemplifies this direction by modeling conditional diversity through Determinantal Point Processes (DPP)\cite{ddp}, but it introduces additional computational overhead. However, the previous three groups of methods largely overlook the varying importance of visual attention heads. They incorrectly assume that all heads contribute equally to interpreting visual information, potentially leading to suboptimal pruning decisions and loss of critical features. Finally, fine-tuning-based methods\cite{fine-tuning, fine-tuning2, similarity} incorporate pruning as a learnable module optimized jointly with model parameters. For example, DART\cite{similarity} adaptively allocates tokens to information-rich regions, achieving higher accuracy but requiring end-to-end training, which increases computational cost and reduces generality.
\section{Method}
In this section, we first provide a brief overview of how MLLMs operate. Next, we formally define the visual token pruning problem and then present our proposed method in detail.
\begin{figure*}[t]
    \centering
    \includegraphics[width=1\textwidth]{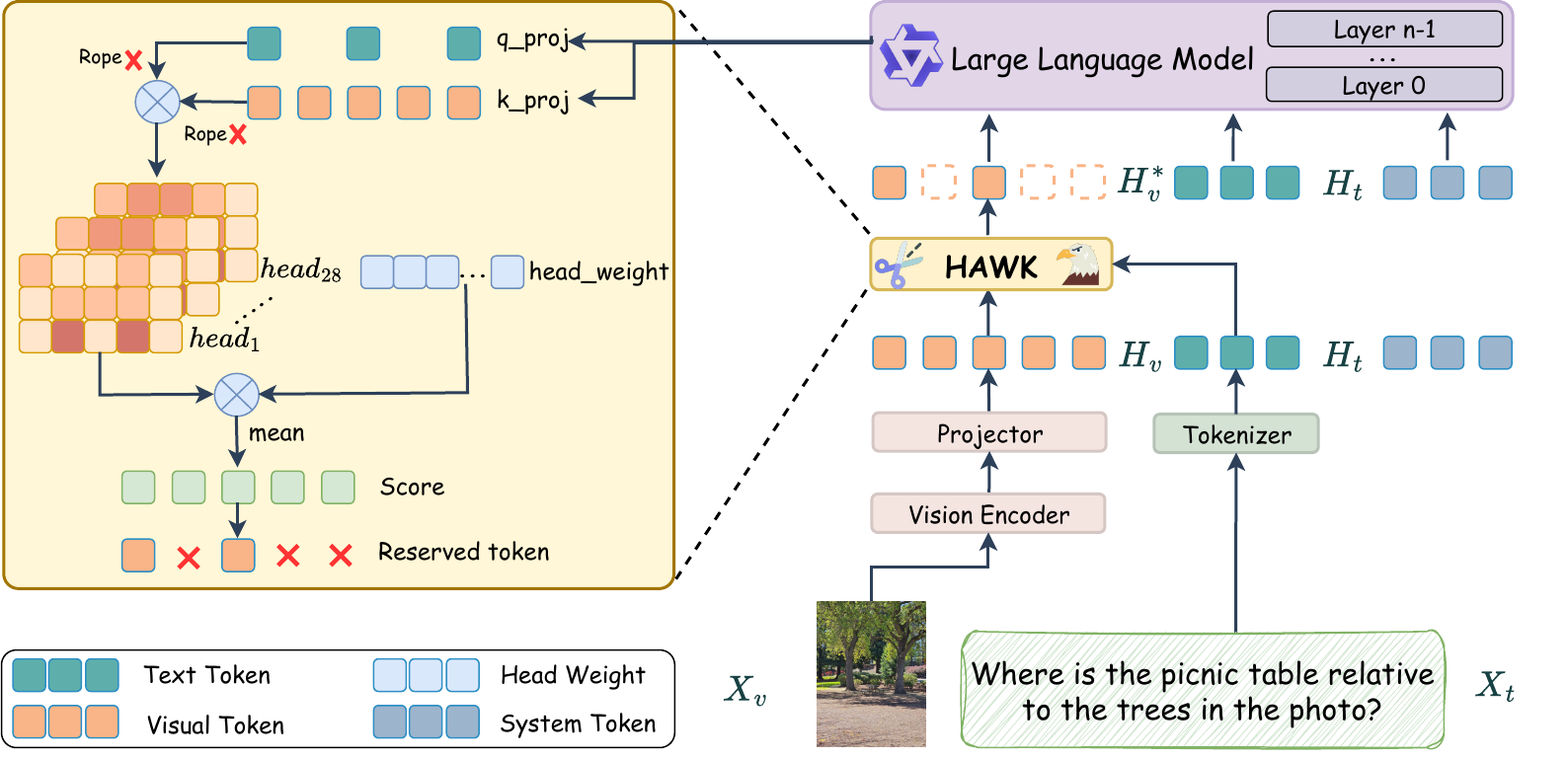}
    \caption{Overview of HAWK. We first employ a text-guided attention mechanism to assess the relevance of each visual embedding relative to the current textual instructions. Then, we integrate it with head importance weights for token pruning. A training-free method that reduces computation with minimal performance loss. }
    \label{fig:method}
\end{figure*}
\subsection{Multimodal Large Language Models}
An MLLM typically takes  a text sequence $\boldsymbol{X}_t$ and a visual sequence $\boldsymbol{X}_v$ (e.g., an image) as input. The text input is first tokenized and embedded into a sequence of textual embeddings denoted by $\boldsymbol{H}_t=\{h_t^1, h_t^2, ...,h_t^N\}$ via a text encoder. Whereas, the visual input is first encoded by a vision encoder and then passed through a projection layer to produce a set of visual token embeddings $\boldsymbol{H}_v=\{h_v^1, h_v^2, ...,h_v^M\}$.
The number of visual tokens often far exceeds that of textual tokens, that is, $M\gg N$.
These two sets of embeddings are then concatenated into a unified input sequence $[H_t; H_v]$ and fed into an LLM. The LLM processes this multimodal sequence auto-regressively to generate an output sequence $\boldsymbol{Y}=\{y_1,y_2,...,y_{O}\}$ according to the following formula:
\begin{equation}
P(\boldsymbol{Y}|\boldsymbol{H}_t,\boldsymbol{H}_v)=\prod_{i=1}^{O}P(y_i|y_{<i},\boldsymbol{H}_t,\boldsymbol{H}_v),
\end{equation}
where each token $y_i$ is predicted based on the previously generated tokens and the fused multimodal context, and $P(\cdot)$ denotes the LLM’s next-token predictive distribution conditioned on the given multimodal input.

\subsection{Visual Token Pruning}
To  mitigate  the computational burden of MLLMs, the basic idea of visual token pruning is to reduce the number of tokens fed into the LLM by selectively discarding some redundant visual tokens, thereby streamlining the input while preserving essential visual information. 
Formally, the objective of visual token pruning is defined as follows:
\begin{equation}
\label{eqaul-2}
    \boldsymbol{\tilde{H}}_v^* = \mathop{\mathrm{argmin}}_{\substack{\tilde{H}_v \subseteq H_v, \\|\tilde{H}_v|= \tilde{M}}}
\mathcal{L}\left(P(\boldsymbol{Y}|\boldsymbol{H}_t,\boldsymbol{H}_v), P(\boldsymbol{Y}|\boldsymbol{H}_t,\boldsymbol{\tilde{H}}_v)\right),
\end{equation}
where $\mathcal{L}(\cdot)$ represents a loss function that measures the difference between the model outputs before and after visual token pruning, and $\tilde{M}$ is the number of visual tokens retained $(\tilde{M}<M)$. 
% The fundamental problem is to discern which visual tokens carry little or no useful information and can therefore be safely removed without compromising model performance.

\subsection{HAWK: Method Overview}
We propose HAWK, a head importance-aware visual token pruning method, as shown in Fig. ~\ref{fig:method}. We reformulate problem~(\ref{eqaul-2}) to retain the visual tokens corresponding to the top-k attention scores. The core idea of HAWK is to jointly decide which visual tokens to retain by integrating static head importance weights with dynamic text-guided attention scores through head importance–aware pruning.

\textbf{Static Visual Head Importance Weights.} We compute head weights using ablation data from several widely used benchmark datasets. Formally, let $N_d$ and $N_h$ denote the total numbers of datasets and attention heads, respectively. For dataset $j$, we define $S_{base,j}$ as the performance score of the baseline model without any ablation, and $S_{i,j}$ (where $i \in \{1, \dots, N_h\}$) as the model’s score after ablating the $i$-th head on that dataset. The performance drop induced by ablating each head is computed as $\Delta S_{i,j} = S_{base,j} - S_{i,j}$. The head importance weight vector $\boldsymbol{W} = \{w_1, \dots, w_{N_h}\}$ is obtained by averaging L1-normalized scores across all datasets:
\begin{equation}
    w_i = \frac{1}{N_d}\sum_{j=1}^{N_d}\frac{S'_{i,j}}{\sum_{i=1}^{N_h} S'_{i,j}},
\end{equation}
where $S'_{i,j} = \Delta S_{i,j} - \min_{i}(\Delta S_{i,j})$ denotes the shifted performance drop of the $i$-th head on the $j$-th dataset, ensuring non-negative values.

\textbf{Dynamic Text-Guided Attention Scores.}
In multimodal tasks, the importance of visual tokens is not fixed, but may vary 
dynamically depending on the input  textual instructions.
%absolute but is closely correlated with the textual embeddings $\boldsymbol{H}_t$. 
To capture such dependency, we introduce a text-guided attention mechanism that dynamically assesses the relevance of each visual embedding $\boldsymbol{H}_v$ and the given textual embeddings $\boldsymbol{H}_t$. 
Specifically, for each attention head $i$, we utilize the query and key projection matrices from the first attention layer of the LLM, denoted by $\boldsymbol{W}_q^i$ and $\boldsymbol{W}_k^i$, to derive the query and key representations as:
%We utilize the query $\boldsymbol{W}_q$ and key $\boldsymbol{W}_k$  projection matrices from the first attention layer of the LLM. for each attention head $i$, we extract $\boldsymbol{W}_q^i$ and $\boldsymbol{W}_k^i$, and derive the query and key representations as:
\begin{equation}
    \boldsymbol{Q}^i = \boldsymbol{H}_t \boldsymbol{W}_q^i, \quad \boldsymbol{K}^i = \boldsymbol{H}_v \boldsymbol{W}_k^i.
\end{equation}
Standard attention typically employs rotary position embedding (RoPE). To eliminate positional bias\cite{endo2025featherthrottlerevisitingvisual} and ensure that the attention depends solely on the semantic correspondence between text and vision, we deliberately  omit the RoPE operation in this step. The position-agnostic attention matrix $\boldsymbol{A}^i \in \mathbb{R}^{N \times M}$ is computed as:
\begin{equation}
    \boldsymbol{A}^i = \frac{(\boldsymbol{Q}^i) (\boldsymbol{K}^i)^T}{\sqrt{d_k}},
\end{equation}
where $d_k$ denotes the key dimension. Each element $A^i_{j,k}$ measures the attention from the $j$-th text token to the $k$-th visual token in head $i$. To obtain the aggregate text-relevance score for each visual token $k$ under head $i$, we average the attention scores over all text tokens, yielding an attention matrix $\boldsymbol{C} \in \mathbb{R}^{N_h \times M}$:
\begin{equation}
    c^i_k = \frac{1}{N} \sum_{j=1}^{N} A^i_{j,k}.
\end{equation}

\textbf{Head Importance-aware Pruning.} HAWK integrates the head weights $\boldsymbol{W}$ with the text-guided attention scores $\boldsymbol{C}$ to evaluate the significance of each visual token. We compute an importance vector $\boldsymbol{I} \in \mathbb{R}^M$, where each element $I_k$ denotes the importance of the $k$-th visual token:
\begin{equation}
    I_k = \sum_{i=1}^{N_h} w_i \cdot c^i_k.
\end{equation}
A higher $I_k$ indicates greater relevance of the $k$-th visual token to the text token. Given a predefined pruning ratio $r$, we retain the top $\tilde{M} = \lfloor M \cdot r \rfloor$ tokens from $\boldsymbol{H}_v$ according to their $I_k$ scores, forming the pruned subset $\boldsymbol{\tilde{H}}_v$. The resulting $\boldsymbol{\tilde{H}}_v$ is concatenated with $\boldsymbol{H}_t$ and passed to the subsequent LLM layers for further processing.

% 颜色
\definecolor{HeadBG}{RGB}{240,240,240}   % 表头灰
\definecolor{SecBG}{RGB}{240,240,240}    % 分段行灰
\definecolor{RowBlue}{RGB}{230,246,255}  % 我们方法（HAWK(Ours)）淡蓝
\definecolor{RowRed}{RGB}{255, 209, 209}   % 对比方法淡红
\definecolor{RowYellow}{RGB}{255, 245, 228}
\definecolor{RowPurple}{RGB}{210, 218, 255}
\definecolor{GoodGreen}{RGB}{0,140,0}    % 绿色注释

% 小工具：绿色“(↓xx%)”注释
\newcommand{\downpct}[1]{\textcolor{GoodGreen}{\small(\textbf{\ensuremath{\downarrow}#1})}}
\begin{table*}[t]
\centering
\caption{{\textbf{Performance comparison of different pruning methods on Qwen2.5-VL-7B using fixed resolution input on image benchmarks.} \textbf{Rel.} represents the average percentage of performance maintained.}\colorbox{RowRed}{Similarity-based method} is shown with a red background, \colorbox{RowYellow}{attention-based method} is with a yellow background, and \colorbox{RowPurple}{similarity\&attention-based method} is with a purple background.}
\label{tab:result_qwen_fixed}
\setlength{\tabcolsep}{5.5pt}
\renewcommand{\arraystretch}{1.2}

\resizebox{\textwidth}{!}{%
\begin{tabular}{lccccccccccc}
\toprule
\textbf{Method} & \textbf{HallBench} & \textbf{MME} & \textbf{TextVQA} & \textbf{ChartQA} & \textbf{AI2D} & \textbf{RealWorldQA} & \textbf{CCBench} & \textbf{OCRVQA} & \textbf{SQA-IMG} &\textbf{POPE}& \textbf{Rel.} \\
\midrule

% ========== Upper Bound ==========
\rowcolor{SecBG}
\multicolumn{12}{c}{\textit{Upper Bound, All 1296 Tokens (100\%)}} \\
Qwen2.5-VL-7B
 & 46.8 & 2322.0 & 85.1 & 86.3 & 80.3 & 66.3 & 58.0 & 70.7 & 72.2&85.3 & 100.0\% \\
\midrule

% ========== Retain 512 tokens ==========
\rowcolor{SecBG}
\multicolumn{12}{c}{\textit{Retain 512 Tokens} \downpct{60.5\%}} \\

\rowcolor{RowRed}
DivPrune (CVPR25)
 & 43.5 & 2258.0 & 81.8 & 80.3 & 78.7 & 63.8 & 55.1 & 69.0 & 71.0 & 84.4&96.4\% \\
\rowcolor{RowYellow} 
FastV (ECCV24)
 & 42.4 & 2318.0 & 84.1 & 82.7 & 78.8 & 64.7 & 55.8 & 67.8 & 71.5 & 84.0&97.0\% \\
\rowcolor{RowPurple}
CDPruner (NeurIPS25)
 & 42.5 & 2327.0 & 84.2 & 82.8 & 78.9 & 64.8 & 55.7 & 67.1 & 71.3 & 82.6&96.8\% \\
\rowcolor{RowBlue}
\textbf{HAWK}(Ours)
 & \textbf{44.1} & \textbf{2342.0} & \textbf{85.3} & \textbf{84.1} & \textbf{79.6} & \textbf{65.4} & \textbf{57.3} & \textbf{71.0} & \textbf{72.6} &\textbf{85.0} &\textbf{99.0\%} \\
\midrule

% ========== Retain 256 tokens ==========
\rowcolor{SecBG}
\multicolumn{12}{c}{\textit{Retain 256 Tokens} \downpct{80.2\%}} \\

\rowcolor{RowRed}
DivPrune (CVPR25)
 & 39.8 & 2162.0 & 75.8 & 69.0 & 76.4 & 60.5 & 53.5 & 65.4 & 70.1 & \textbf{83.5}&91.3\% \\
\rowcolor{RowYellow} 
FastV (ECCV24)
 & 39.0 & 2233.0 & 81.5 & 71.4 & 76.2 & 62.5 & 54.1 & 65.6 & 70.9 & 81.2&92.6\% \\
\rowcolor{RowPurple}
CDPruner (NeurIPS25)
 & 40.1 & 2245.0 & 82.4 & 73.0 & 77.5 & 62.2 & 54.3 & 61.0 & 70.7 & 79.4&92.5\% \\
\rowcolor{RowBlue}
\textbf{HAWK}(Ours)
 & \textbf{42.4} & \textbf{2245.0} & \textbf{83.7} & \textbf{77.2} & \textbf{77.7} & \textbf{63.7} & \textbf{55.5} & \textbf{70.3} & \textbf{71.9} & 82.9&\textbf{96.0\%} \\
\midrule

% ========== Retain 128 tokens ==========
\rowcolor{SecBG}
\multicolumn{12}{c}{\textit{Retain 128 Tokens} \downpct{90.1\%}} \\

\rowcolor{RowRed}
DivPrune (CVPR25)
 & 33.4 & 1979.0 & 66.6 & 52.5 & 72.1 & 57.9 & 48.4 & 57.8 & 68.6 & \textbf{81.8}&82.8\% \\
\rowcolor{RowYellow}
FastV (ECCV24)
 & 33.8 & 2028.0 & 73.8 & 54.3 & 71.4 & 57.8 & 46.7 & 62.7 & 69.4 & 75.8&83.9\% \\
\rowcolor{RowPurple}
CDPruner (NeurIPS25)
 & 37.2 & 2127.0 & 77.8 & 59.2 & 74.0 & 59.1 & 49.4 & 54.3 & 69.3 & 75.6&85.9\% \\
\rowcolor{RowBlue}
\textbf{HAWK}(Ours)
 & \textbf{40.5} & \textbf{2137.0} & \textbf{79.2} & \textbf{64.4} & \textbf{75.0} & \textbf{59.4} & \textbf{49.4} & \textbf{67.8} & \textbf{70.2} & 77.8&\textbf{90.0\%} \\
\bottomrule
\end{tabular}
}% end resizebox
\end{table*}
\section{Experiments}
In this section, we present a comprehensive set of experiments to evaluate the effectiveness of our method against prior approaches across diverse models, tasks, and datasets. Moreover, we conduct ablation studies to assess the contribution of each component and validate the model’s design, and further analyze inference efficiency, demonstrating substantial gains in end-to-end latency and overall computational performance.
% We first conduct extensive evaluations on advanced MLLMs, including Qwen2.5-VL-7B and InternVL3-8B, over multiple vision-language benchmarks, demonstrating the superiority of our approach in maintaining high task accuracy compared with state-of-the-art methods. 
% We then perform detailed ablation studies to analyze the independent contributions of key components and to validate the design rationale of our model. Finally, we provide an in-depth analysis of inference efficiency, showing that our method achieves significant acceleration in end-to-end latency and other efficiency metrics.

\subsection{Experimental Setup}

\textbf{Evaluation benchmarks:} 
We select ten image-based multimodal benchmarks for extensive evaluation. We divide these benchmarks into two main categories: the first category consists of six general and reasoning VQA tasks, including POPE\cite{pope}, HallBench\cite{hallbench}, MME\cite{mme}, ScienceQA-IMG\cite{scienceqa}, RealWorldQA\cite{realworldqa}, and CCBench\cite{ccbench}. The second category comprises four text- and diagram-oriented VQA tasks, including TextVQA\cite{textvqa}, OCRVQA\cite{ocrvqa}, ChartQA\cite{chartqa}, and AI2D\cite{ai2d}. We also conduct experiments on two video understanding benchmarks, including Video-MME\cite{videomme} and WorldSense\cite{worldsense}. All experiments on these benchmarks follow their default settings and evaluation metrics.

\textbf{Comparison methods and models:} 
We select three mainstream plug-and-play pruning methods as comparative baselines, each adopting a different technical strategy. These baselines include the attention-based FastV\cite{FastV}, similarity-based DivPrune\cite{divprune}, and the hybrid method CDPruner\cite{cdpruner}, which combines both importance and similarity. We conducted our experiments on the current state-of-the-art open-source models, Qwen2.5-VL\cite{qwen} and InternVL3\cite{Internvl}.

\textbf{Experimental environment:}
We conduct all benchmark evaluations on 8 GPUs (96GB VRAM) using the VLMEvalkit\cite{vlmevalkit}, while we perform the efficiency analysis on a single  GPU.

\begin{table*}[t]
\centering
\caption{{\textbf{Performance comparison of different pruning methods on InternVL3-8B using fixed resolution input on image benchmarks.} \textbf{Rel.} represents the average percentage of performance maintained.} \colorbox{RowRed}{Similarity-based method} is shown with a red background,  and \colorbox{RowPurple}{similarity\&attention-based method} is with a purple background.}
%\colorbox{RowYellow}{attention-based methods} with a yellow background
\label{tab:result_intervlanyres}
\setlength{\tabcolsep}{5.5pt}
\renewcommand{\arraystretch}{1.2}

\resizebox{\textwidth}{!}{%
\begin{tabular}{lccccccccccc}
\toprule
\textbf{Method} & \textbf{HallBench} & \textbf{MME} & \textbf{TextVQA} & \textbf{ChartQA} & \textbf{AI2D} & \textbf{RealWorldQA} & \textbf{CCBench} & \textbf{OCRVQA} & \textbf{SQA-IMG} &\textbf{POPE}& \textbf{Rel.} \\
\midrule

% ========== Upper Bound ==========
\rowcolor{SecBG}
\multicolumn{12}{c}{\textit{Upper Bound,All 1280 Tokens (100\%)}} \\
InternVL3-8B
 & 49.8 & 2404.0 & 79.9 & 80.4 & 83.4 & 68.8 & 70.2 & 41.1 & 96.7&90.7 & 100.0\% \\
\midrule

% ========== 60% ==========
\rowcolor{SecBG}
\multicolumn{12}{c}{\textit{Retain 512 Tokens } \downpct{60\%}} \\

\rowcolor{RowRed}
DivPrune (CVPR25)
 & 43.7 & 2226.0 & 74.9 & 69.4 & 81.3 & 64.4 & 68.4 & 37.7 & 87.3 & 90.8&93.0\% \\
% \rowcolor{RowYellow} 
% FastV (ECCV24)
%  & 42.5 & 2283.0 & 84.1 & 82.5 & 78.2 & 64.7 & 55.9 & 67.8 & 71.6 & 84.1&96.1\% \\
\rowcolor{RowPurple}
CDPruner (NeurIPS25)
 & 37.7 & 2136.0 & 55.8 & 52.9 & 76.6 & 59.5 & 66.9 & 33.3 & 84.5 & 88.6&84.0\% \\
\rowcolor{RowBlue}
\textbf{HAWK}(Ours)
 & \textbf{48.5} & \textbf{2421.0} & \textbf{79.3} & \textbf{76.5} & \textbf{82.6} & \textbf{67.2} & \textbf{69.8} & \textbf{39.5} & \textbf{95.2} &\textbf{90.8} &\textbf{98.2\%} \\
\midrule

% ========== 80% ==========
\rowcolor{SecBG}
\multicolumn{12}{c}{\textit{Retain 256 Tokens} \downpct{80\%}} \\

\rowcolor{RowRed}
DivPrune (CVPR25)
 & 39.2 & 2216.0 & 67.9 & 55.0 & 77.5 & 62.4 & 63.9 & 35.5 & 84.3 & 89.8&87.1\% \\
% \rowcolor{RowYellow} 
% FastV (ECCV24)
%  & 38.2 & 2236.0 & 81.9 & 72.3 & 76.4 & 63.1 & 54.5 & 66.1 & 71.2 & 81.8&92.3\% \\
\rowcolor{RowPurple}
CDPruner (NeurIPS25)
 & 34.9 & 1887.0 & 41.8& 37.9 & 69.5 & 55.9 & 63.7 & 29.6 & 76.5 & 85.9&74.8\% \\
\rowcolor{RowBlue}
\textbf{HAWK}(Ours)
 & \textbf{45.0} & \textbf{2341.0} & \textbf{76.5} & \textbf{66.1} & \textbf{79.1} & \textbf{65.0} & \textbf{70.2} & \textbf{38.0} & \textbf{91.8} & \textbf{90.0}&\textbf{94.1\%} \\
\midrule

% % ========== 90% ==========
% \rowcolor{SecBG}
% \multicolumn{12}{c}{\textit{Pruning Ratio} \downpct{90\%}} \\

% \rowcolor{RowRed}
% DivPrune (CVPR25)
%  & 33.9 & 2047.0 & 67.4 & 51.7 & 73.1 & 57.7 & \textbf{50.7} & 59.0 & 68.9 & \textbf{81.5}&83.3\% \\
% \rowcolor{RowYellow}
% FastV (ECCV24)
%  & 33.6 & 2022.0 & 72.2 & 57.6 & 73.2 & 57.5 & 47.0 & 62.5 & 69.4 & 76.2&83.7\% \\
% \rowcolor{RowPurple}
% CDPruner (NeurIPS25)
%  & 28.8 & 1970.0 & 57.2 & 28.6 & 75.5 & 60.3 & 54.3 & 53.6& 69.4 & 75.7&77.9\% \\
% \rowcolor{RowBlue}
% \textbf{HAWK}(Ours)
%  & \textbf{40} & \textbf{2101.0} & \textbf{79.8} & \textbf{65.2} & \textbf{75.3} & \textbf{60.4} & 49.8 & \textbf{68.4} & \textbf{71.6} & 77.9&\textbf{89.7\%} \\
% \bottomrule
\end{tabular}
}% end resizebox
\end{table*}

\begin{table*}[t]
\centering
\caption{\textbf{Performance comparison of different pruning methods on Qwen2.5-VL-7B using native resolution input on image benchmarks.} \textbf{Rel.} represents the average percentage of performance maintained.\colorbox{RowRed}{Similarity-based method} is shown with a red background, \colorbox{RowYellow}{attention-based method} is with a yellow background, and \colorbox{RowPurple}{similarity\&attention-based method} is with a purple background.}
\label{tab:result_qwen_nativeres}
\setlength{\tabcolsep}{5.5pt}
\renewcommand{\arraystretch}{1.2}

\resizebox{\textwidth}{!}{%
\begin{tabular}{lccccccccccc}
\toprule
\textbf{Method} & \textbf{HallBench} & \textbf{MME} & \textbf{TextVQA} & \textbf{ChartQA} & \textbf{AI2D} & \textbf{RealWorldQA} & \textbf{CCBench} & \textbf{OCRVQA} & \textbf{SQA-IMG} &\textbf{POPE}& \textbf{Rel.} \\
\midrule

% ========== Upper Bound ==========
\rowcolor{SecBG}
\multicolumn{12}{c}{\textit{Native-Resolution (100\%)}} \\
Qwen2.5-VL-7B
 & 46.5 & 2315.0 & 85.2 & 86.2 & 80.7 & 67.7 & 59.8 & 71 & 72.8&86.5 & 100.0\% \\
\midrule

% ========== 60% ==========
\rowcolor{SecBG}
\multicolumn{12}{c}{\textit{Pruning Ratio} \downpct{60\%}} \\

\rowcolor{RowRed}
DivPrune (CVPR25)
 & 45.8 & 2274.0 & 82.7 & 80.6 & 78.6 & 64.4 & 56.4 & 69.8 & 71.6 & 84.6&96.9\% \\
\rowcolor{RowYellow} 
FastV (ECCV24)
 & 42.5 & 2283.0 & 84.1 & 82.5 & 78.2 & 64.7 & 55.9 & 67.8 & 71.6 & 84.1&96.1\% \\
\rowcolor{RowPurple}
CDPruner (NeurIPS25)
 & 39.3 & 2248.0 & 77.8 & 65.4 & 78.9 & 65.8 & 56.0 & 67.3 & 72.0 & 82.8&92.7\% \\
\rowcolor{RowBlue}
\textbf{HAWK}(Ours)
 & \textbf{46.5} & \textbf{2313.0} & \textbf{85.0} & \textbf{83.6} & \textbf{79.9} & \textbf{67.6} & \textbf{59.8} & \textbf{71.0} & \textbf{73.6} &\textbf{85.8} &\textbf{99.6\%} \\
\midrule

% ========== 80% ==========
\rowcolor{SecBG}
\multicolumn{12}{c}{\textit{Pruning Ratio} \downpct{80\%}} \\

\rowcolor{RowRed}
DivPrune (CVPR25)
 & 39.0 & 2196.0 & 76.8 & 69.0 & 77.5 & 62.7 & 53.7 & 66.8 & 71.3 & 83.4&91.6\% \\
\rowcolor{RowYellow} 
FastV (ECCV24)
 & 38.2 & 2236.0 & 81.9 & 72.3 & 76.4 & 63.1 & 54.5 & 66.1 & 71.2 & 81.8&92.3\% \\
\rowcolor{RowPurple}
CDPruner (NeurIPS25)
 & 34.5 & 2166.0 & 68.8& 42.6 & 78.1 & 64.2 & 55.2 & 61.4 & 70.8 & 80.2&85.8\% \\
\rowcolor{RowBlue}
\textbf{HAWK}(Ours)
 & \textbf{42.8} & \textbf{2311.0} & \textbf{83.0} & \textbf{76.8} & \textbf{78.1} & \textbf{65.0} & \textbf{56.7} & \textbf{70.5} & \textbf{73.2} & \textbf{83.5}&\textbf{96.2\%} \\
\midrule

% ========== 90% ==========
\rowcolor{SecBG}
\multicolumn{12}{c}{\textit{Pruning Ratio} \downpct{90\%}} \\

\rowcolor{RowRed}
DivPrune (CVPR25)
 & 33.9 & 2047.0 & 67.4 & 51.7 & 73.1 & 57.7 & \textbf{50.7} & 59.0 & 68.9 & \textbf{81.5}&83.3\% \\
\rowcolor{RowYellow}
FastV (ECCV24)
 & 33.6 & 2022.0 & 72.2 & 57.6 & 73.2 & 57.5 & 47.0 & 62.5 & 69.4 & 76.2&83.7\% \\
\rowcolor{RowPurple}
CDPruner (NeurIPS25)
 & 28.8 & 1970.0 & 57.2 & 28.6 & 75.5 & 60.3 & 54.3 & 53.6& 69.4 & 75.7&77.9\% \\
\rowcolor{RowBlue}
\textbf{HAWK}(Ours)
 & \textbf{40} & \textbf{2101.0} & \textbf{79.8} & \textbf{65.2} & \textbf{75.3} & \textbf{60.4} & 49.8 & \textbf{68.4} & \textbf{71.6} & 77.9&\textbf{89.7\%} \\
\bottomrule
\end{tabular}
}% end resizebox
\end{table*}

\begin{table*}[t]
\centering
\caption{{\textbf{Performance comparison of different pruning methods on Qwen2.5-VL-7B using native resolution input on video benchmarks.} \textbf{Rel.} represents the average percentage of performance maintained.} \colorbox{RowRed}{Similarity-based method} is shown with a red background, \colorbox{RowYellow}{attention-based method} is with a yellow background, and \colorbox{RowPurple}{similarity\&attention-based method} is with a purple background.}
\label{tab:result_video}
\setlength{\tabcolsep}{4.5pt} % 稍微调整列间距以适应更多列
\renewcommand{\arraystretch}{1.2}

\resizebox{\textwidth}{!}{%
\begin{tabular}{lcccccccccccc}
\toprule
\textbf{Method} & \multicolumn{4}{c}{\textbf{VideoMME}} & \multicolumn{7}{c}{\textbf{WorldSense}} & \multirow{2}{*}{\textbf{Rel.}} \\
\cmidrule(lr){2-5} \cmidrule(lr){6-12}
& \textbf{Short} & \textbf{Medium} & \textbf{Long} & \textbf{Overall} & \textbf{$<1$min} & \textbf{$1\sim2$min} & \textbf{$2\sim4$min} & \textbf{$4\sim6$min} & \textbf{$6\sim8$min} & \textbf{$>8$min} & \textbf{Overall} & \\
\midrule

% ========== Upper Bound ==========
\rowcolor{SecBG}
\multicolumn{13}{c}{\textit{Native-Resolution (100\%)}} \\
Qwen2.5-VL-7B
& 61.9 & 51.2 & 46.4 & 53.2  % VideoMME
& 38.7 & 34.9 & 30.7 & 35.3 & 31.6 & 33.6 & 35.1  % WorldSense
& 100\% \\ 
\midrule

% ========== 60% ==========
\rowcolor{SecBG}
\multicolumn{13}{c}{\textit{Pruning Ratio} \downpct{60\%}} \\
\midrule

\rowcolor{RowRed}
DivPrune (CVPR25)
& 61.0 & 49.8 & 44.6 & 51.8 & 37.9 & 33.6 & 31.0 & 33.0 & 33.8 & 30.2 & 34.3 & 97.5\% \\
\rowcolor{RowYellow} 
FastV (ECCV24)
& 59.8 & 51.7 & 45.8 & 52.4 & 37.1 & 35.3 & 31.4 & 34.4 & 27.2 & 28.9 & 34.4 & 98.3\% \\
\rowcolor{RowPurple}
CDPruner (NeurIPS25)
& 59.8 & 51.9 & 44.6 & 52.1 & 34.9 & 33.4 & 28.3 & 30.8 & 33.8 &32.2 & 32.6 & 95.4\% \\
\rowcolor{RowBlue}
\textbf{HAWK}(Ours)
& 61.3 & 50.8 & 45.9 & \textbf{52.7} & 37.9& 34.3& 31.9 & 34.4 & 30.1 & 30.9 & \textbf{34.6} & \textbf{98.8\%} \\
\midrule

% ========== 80% ==========
\rowcolor{SecBG}
\multicolumn{13}{c}{\textit{Pruning Ratio} \downpct{80\%}} \\
\midrule

\rowcolor{RowRed}
DivPrune (CVPR25)
& 58.7 & 50.6 & 43.9 & 51.0 & 36.9 & 33.1 & 30.6 & 33.0 & 30.9 & 28.2 & \textbf{33.5} & 95.7\% \\
\rowcolor{RowYellow} 
FastV (ECCV24)
& 57.4 & 49.6 & 45.6 & 50.9 & 35.3 & 33.3 & 30.4 & 35.3 & 32.4 & 30.2 & 33.3 & 95.3\% \\
\rowcolor{RowPurple}
CDPruner (NeurIPS25)
& 58.8 & 50.6 & 44.0 & 51.1 & 36.0 & 34.2 & 28.6 & 33.5 & 30.1 & 30.9 & 33.2 & 95.3\% \\
\rowcolor{RowBlue}
\textbf{HAWK}(Ours)
& 60.1 & 51.0 & 45.6 & \textbf{52.2}& 36.1& 33.5 & 30.1 & 31.7 & 29.4& 33.6 & 33.4 & \textbf{96.6\%} \\
\midrule

% ========== 90% ==========
\rowcolor{SecBG}
\multicolumn{13}{c}{\textit{Pruning Ratio} \downpct{90\%}} \\
\midrule

\rowcolor{RowRed}
DivPrune (CVPR25)
& 57.1 & 50.3 & 42.7 & 50.0 & 34.4 & 33.8 & 30.7 & 30.3 & 31.6 & 30.9 & \textbf{32.8} & 93.7\% \\
\rowcolor{RowYellow}
FastV (ECCV24)
& 55.9 & 48.9 & 45.3 & 50.0 & 33.8 & 32.6 & 29.4 & 32.6 & 27.9 & 32.9 & 32.1 & 92.7\% \\
\rowcolor{RowPurple}
CDPruner (NeurIPS25)
& 57.2 & 49.9 & 44.2 & \textbf{50.4} & 34.4 & 33.1 & 29.5 & 32.1 & 30.1 & 28.9 & 32.4 & 93.5\% \\
\rowcolor{RowBlue}
\textbf{HAWK}(Ours)
& 56.3 & 48.7 & 45.2& 50.1 & 34.6 & 33.5 & 30.1& 30.8 & 30.9 & 30.9 & 32.7 & \textbf{93.7\%} \\
\bottomrule
\end{tabular}
}% end resizebox
\end{table*}
\subsection{HAWK for Fixed-Resolution Input}
We first evaluated HAWK at a fixed resolution on the Qwen2.5VL-7B model. The experimental setup employed a fixed input resolution of 1008$\times$1008, where the baseline number of visual tokens generated by the model was 1296. Based on these settings, we define three pruning strengths, corresponding to retaining 512 (60.5\% pruning ratio), 256 (80.2\% pruning ratio), and 128 (90.1\% pruning ratio) visual tokens. 
As shown in Table \ref{tab:result_qwen_fixed}, HAWK demonstrates outstanding performance across all pruning ratios. Specifically, at a 60.5\% pruning ratio, HAWK maintains 99.0\% of its original average performance, achieving a significant \textbf{2.1 percentage} point improvement over the best-performing baseline method (FastV, 97.0\%). When the pruning ratio increased to 80.2\%, HAWK's average performance remained at 96.0\%, surpassing the second-best method by \textbf{3.4 percentage} points. Even at an extreme pruning ratio of 90.1\%, HAWK still achieves 90.0\% performance, surpassing the best performance of other mainstream methods by \textbf{4.1 percentage} points. These results clearly demonstrate that compared to existing methods, HAWK can more effectively preserve the model's original performance while significantly compressing visual tokens, showcasing its SOTA capabilities. %% 

To further validate the generalizability of HAWK, we migrated it to other advanced open-source MLLM architectures, InternVL3-8B \cite{Internvl}, for additional experiments. This experimental setup employed a fixed input resolution of 896$\times$896, where the baseline number of visual tokens generated by the model was 1280. Based on this, we defined two pruning ratios, corresponding to retaining 512 (60\% pruning ratio) and 256 (80\% pruning ratio) visual tokens. We selected two advanced methods, DivPrune and CDPruner, as comparative baselines.
As shown in Table \ref{tab:result_intervlanyres}, HAWK demonstrates outstanding performance at both pruning ratios. Specifically, at a 60\% pruning ratio, HAWK maintains 98.2\% of its original performance, achieving a significant \textbf{5.2 percentage} point improvement over the other baseline methods. When the pruning ratio is increased to 80\%, HAWK's performance remains at 94.1\%, surpassing the other methods by \textbf{7.0 percentage} points. These results clearly demonstrate that the HAWK framework possesses excellent generalization capabilities, enabling it to efficiently compress visual tokens while maintaining top-tier performance across different model architectures.

\subsection{HAWK for Native-Resolution Input}
The Qwen2.5VL-7B model is capable of processing image inputs at various resolutions. Such dynamic and variable input length imposes stricter requirements on the robustness of pruning methods. Therefore, beyond the fixed-resolution benchmarks, we further evaluated HAWK's generalization performance in native-resolution scenarios. We adopted pruning strengths corresponding to the preceding experiments (60.5\%, 80.2\%, 90.1\%), setting the target pruning ratios to 60\%, 80\%, and 90\%, respectively.
As shown in Table \ref{tab:result_qwen_nativeres}, HAWK demonstrates leading performance across all pruning ratios. Specifically, at a 60\% pruning ratio, HAWK maintains 99.6\% of its original performance, achieving a significant \textbf{2.7 percentage} point improvement over other methods. When the pruning ratio is increased to 80\%, HAWK's performance remains at 96.2\%, surpassing other methods by \textbf{3.9 percentage} points. Even at an extreme pruning ratio of 90\%, HAWK still achieves 89.7\% performance, outperforming other methods by \textbf{6.0 percentage} points. These results strongly demonstrate that HAWK not only performs exceptionally on controlled benchmarks but also possesses excellent robustness when processing dynamic and variable inputs.

\begin{table*}[t] 
\centering
\caption{\textbf{Comparison results of HAWK and baselines on Qwen2.5-VL-7B across MME benchmark}. Original denotes the original model without pruning. The computation of the KV cache is performed on individual samples. The memory refers to the GPU memory consumption during model inference. Since FastV does not support FlashAttention, it is excluded from the comparison.}
\label{tab:pruning_comparison_wide_v3} % 建议使用新标签
\small
\setlength{\tabcolsep}{5.5pt}
% 9列表格结构：l (Method) + 4列r (0.6) + 4列r (0.8)
\begin{tabular}{@{}l rccc rccc @{}} 
\toprule
% 1. 表头：Ratio 分组
& \multicolumn{4}{c}{\textbf{Ratio 0.6}} & \multicolumn{4}{c}{\textbf{Ratio 0.8}} \\
\cmidrule(lr){2-5} \cmidrule(lr){6-9} % 在多列标题下画线
% 2. 表头：具体列名
\textbf{Method} & \textbf{Score} & \textbf{E2E Latency} & \textbf{KV Cache (MB)} & \textbf{Mem. (GB)} & \textbf{Score} & \textbf{E2E Latency} & \textbf{KV Cache (MB)} & \textbf{Mem. (GB)} \\
\midrule
% 3. Original 基线行：数据在两侧重复，作为共同的基线
Original        & 2315.0 & 20m 15s & 668 & 16.9 & 2315.0 & 20m 15s & 668 & 16.9 \\
\midrule % 4. 分隔线：将基线与剪枝方法分开
\textbf{HAWK}   & \textbf{2313.0} & \textbf{16m 10s (x1.25)} & \textbf{276} & \textbf{16.1} & \textbf{2311.0} & \textbf{15m 04s (x1.34)} & \textbf{148} & \textbf{15.7} \\
DivPrune        & 2274.0 & 18m 46s (x1.08) & 276 & 16.1 & 2196.0 & 15m 31s (x1.31) & 148 & 15.7 \\
CDPruner        & 2248.0 & 19m 59s (x1.01)& 276 & 16.1 & 2166.0 & 16m 33s (x1.22)& 148 & 15.7 \\
\bottomrule
\end{tabular}
\end{table*}
\subsection{HAWK for Video Understanding}
Compared to image understanding tasks, video understanding tasks inherently involve a significantly larger number of visual tokens, making visual token pruning a source of greater potential benefit in terms of efficiency. To validate the effectiveness of the HAWK method in multi-modal large model video understanding tasks, we conducted experiments on the Qwen2.5-VL-7B architecture, selecting two distinct video benchmarks, VideoMME and WorldSense, which feature videos of different lengths. Token pruning was performed using native resolution video inputs, standardized to 8 frames, and evaluated across three pruning ratios: 0.6, 0.8, and 0.9. Our comparison included established methods such as FastV, DivPrune, and CDPruner. The results, as detailed in Table \ref{tab:result_video}, show that HAWK achieved the leading performance among the four methods: at a 60\% pruning ratio, HAWK maintained 98.8\% of the original performance; at an 80\% pruning ratio, it preserved 96.6\% of the original performance; and even at a 90\% pruning ratio, HAWK still maintained 93.7\% of the original performance, consistently leading all four compared methods. These findings conclusively demonstrate the effectiveness of HAWK for efficient video understanding within multi-modal large language models.

\subsection{Efficiency Analysis}
To evaluate the effectiveness of HAWK, we conducted a comprehensive comparison with existing pruning baselines under two pruning ratios (0.6 and 0.8) on Qwen2.5-VL-7B across the MME benchmark. As shown in Table \ref{tab:pruning_comparison_wide_v3}, HAWK achieves the best trade-off between performance and efficiency. At a pruning ratio of 0.6, it attains an MME score of 2313.0, which is almost identical to the original model, while reducing E2E latency from 20m 15s to 16m 10s (\textbf{×1.25} speedup). When the pruning ratio increases to 0.8, HAWK maintains comparable accuracy with a score of 2311.0 and further accelerates inference to 15m 04s (\textbf{×1.34} speedup), outperforming both DivPrune and CDPruner in latency reduction. In addition, HAWK notably decreases the KV cache and the memory usage, requiring only 276 MB and 16.1 GB at ratio 0.6, and 148 MB and 15.7 GB at ratio 0.8. These results demonstrate that HAWK delivers superior inference efficiency with minimal performance degradation, validating its effectiveness as a practical pruning strategy for MLLMs.

% \begin{table}[t]
% \centering
% \caption{Comparison results of HAWK and baselines on Qwen2.5-VL-7B across MME datasets.  Original denotes the original model without pruning.}
% \label{tab:pruning_comparison}
% \small
% \setlength{\tabcolsep}{5.5pt}
% \begin{tabular}{@{}ll r r r r @{}} 
% \toprule
% \textbf{Ratio} & \textbf{Method} & \textbf{Score} & \textbf{E2E Latency} &\textbf{kvcache(MB)}& \textbf{Mem. (GB)} \\
% \midrule
% Original & -         & 2315.0 & 20m 15s &668& 16.9 \\
% \midrule
% \multirow{3}{*}{0.6}
%   & \textbf{HAWK}     & \textbf{2313.0} & \textbf{16m 10s} & \textbf{276}&16.1 \\
%   & DivPrune  & 2274.0 & 18m 46s & 276&16.1 \\
%   & CDPruner  & 2248.0 & 19m 59s & 276&16.1 \\
% \midrule
% \multirow{3}{*}{0.8}
%   & \textbf{HAWK}     & \textbf{2311.0} & \textbf{15m 04s} & \textbf{148}&15.7 \\
%   & DivPrune  & 2196.0 & 15m 31s &148& 15.7 \\
%   & CDPruner  & 2166.0 & 16m 33s & 148&15.7 \\
% \bottomrule
% \end{tabular}
% \end{table}
% 请确保在导言区加载了 \usepackage{booktabs}

\subsection{Ablation Study}
\begin{figure}[t]
    \centering
    % 子图 1
    \begin{subfigure}[b]{0.48\linewidth}
        \centering
        \includegraphics[width=\linewidth]{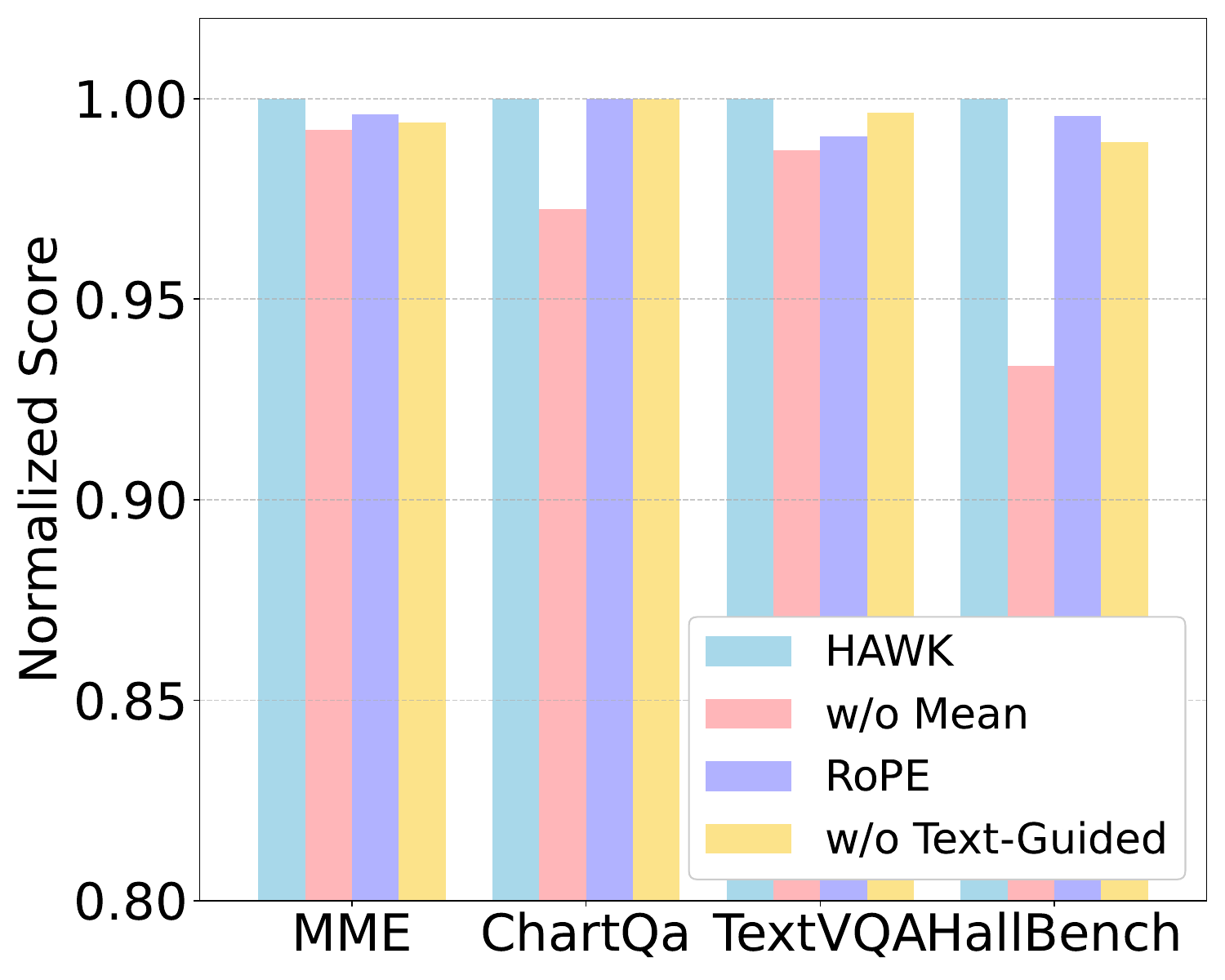}
        \caption{Pruning Ratio 0.6}
        \label{fig:ablation_06}
    \end{subfigure}
    \hfill
    % 子图 2
    \begin{subfigure}[b]{0.48\linewidth}
        \centering
        \includegraphics[width=\linewidth]{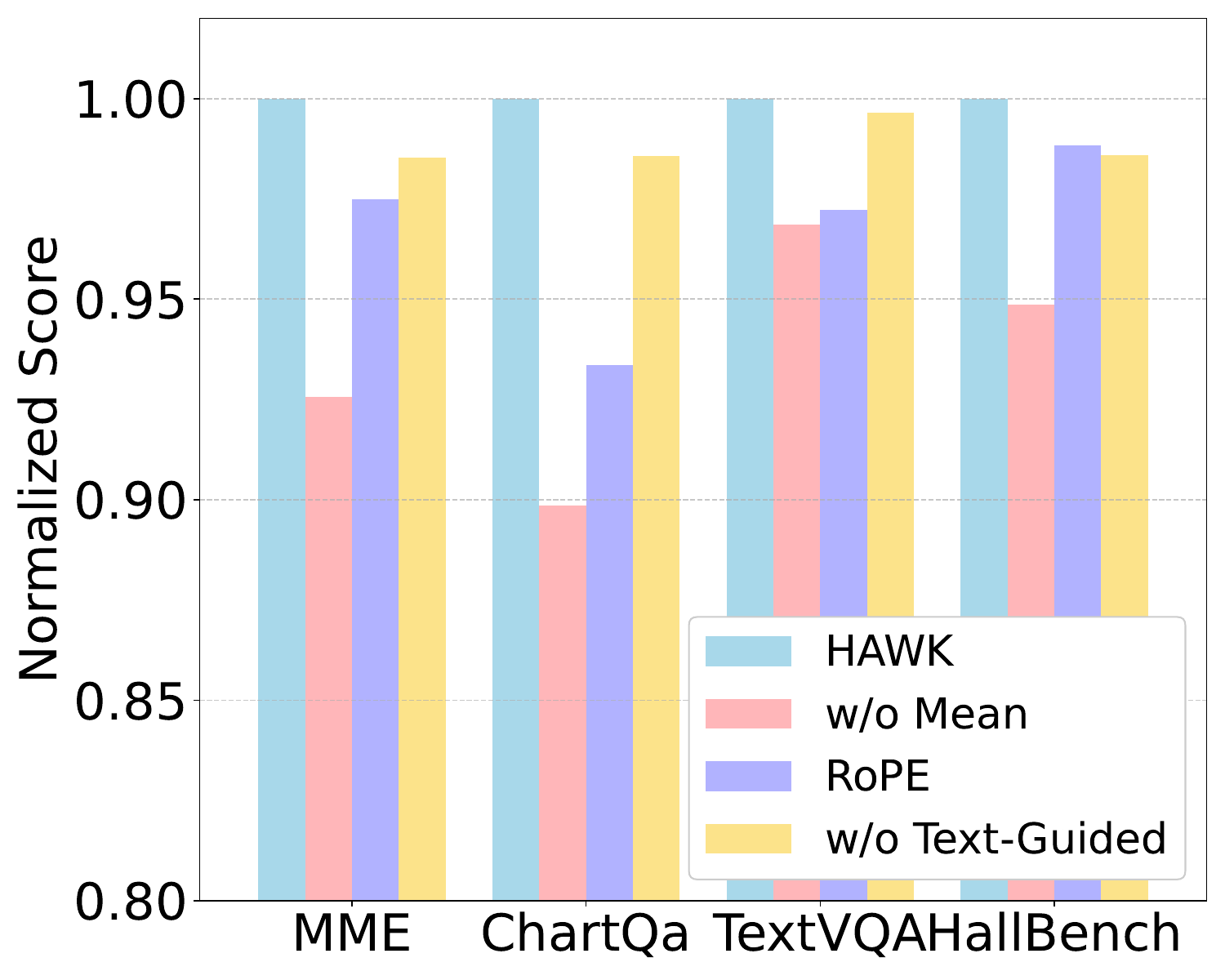}
        \caption{Pruning Ratio 0.8}
        \label{fig:ablation_08}
    \end{subfigure}

    \caption{Ablation study results under different pruning ratios.}
    \label{fig:ablation}
\end{figure}

To validate the effectiveness of HAWK's core designs, we conducted relevant ablation studies on the Qwen2.5-VL-7B model using native resolution inputs. The experiments were performed at two pruning ratios (0.6 and 0.8) across four key benchmarks. To isolate and evaluate the impact of HAWK's individual core components, we established three critical control groups. First, to verify the necessity of the head importance-aware mechanism, we tested a variant that utilized averaged head attention. Second, to assess the optimality of our precise text-guided query strategy, we evaluated a scheme using other tokens following the text sequence as the query. Finally, we examined the impact of reintroducing positional encoding on performance. As   illustrated in Fig. \ref{fig:ablation}, the experimental results clearly demonstrate that all these control variants led to varying degrees of performance degradation. This strongly confirms that each of HAWK's core designs, namely its head importance weighting mechanism, precise text-guidance strategy, and specific handling of positional information, is indispensable for achieving its superior performance.

\section{Conclusion}
In this work, we propose HAWK, a head importance-aware visual token pruning framework for MLLMs.HAWK reveals that attention heads contribute unequally to visual understanding and leverages this insight to guide token pruning. By integrating head importance weights with text-guided attention, HAWK selectively preserves task-critical visual tokens while discarding redundant ones. The method is entirely training-free and applicable to diverse MLLMs. Extensive experiments demonstrate that HAWK achieves state-of-the-art performance, retaining 96.0\% of the original accuracy after pruning 80.2\% of visual tokens, while reducing end-to-end latency to 74.4\% and lowering GPU memory consumption. These results highlight the effectiveness and practicality of HAWK for efficient multimodal inference.

{
    \small
    \bibliographystyle{ieeenat_fullname}
    \bibliography{main}
}
\clearpage
\setcounter{page}{1}
\maketitlesupplementary
\appendix
\section{Extended Analysis on Visual Head Ablation}
\label{sec:rationale}
\begin{figure}[ht]
  \centering
  % 请确保文件名与生成的PDF文件名一致
  \includegraphics[width=0.7\linewidth]{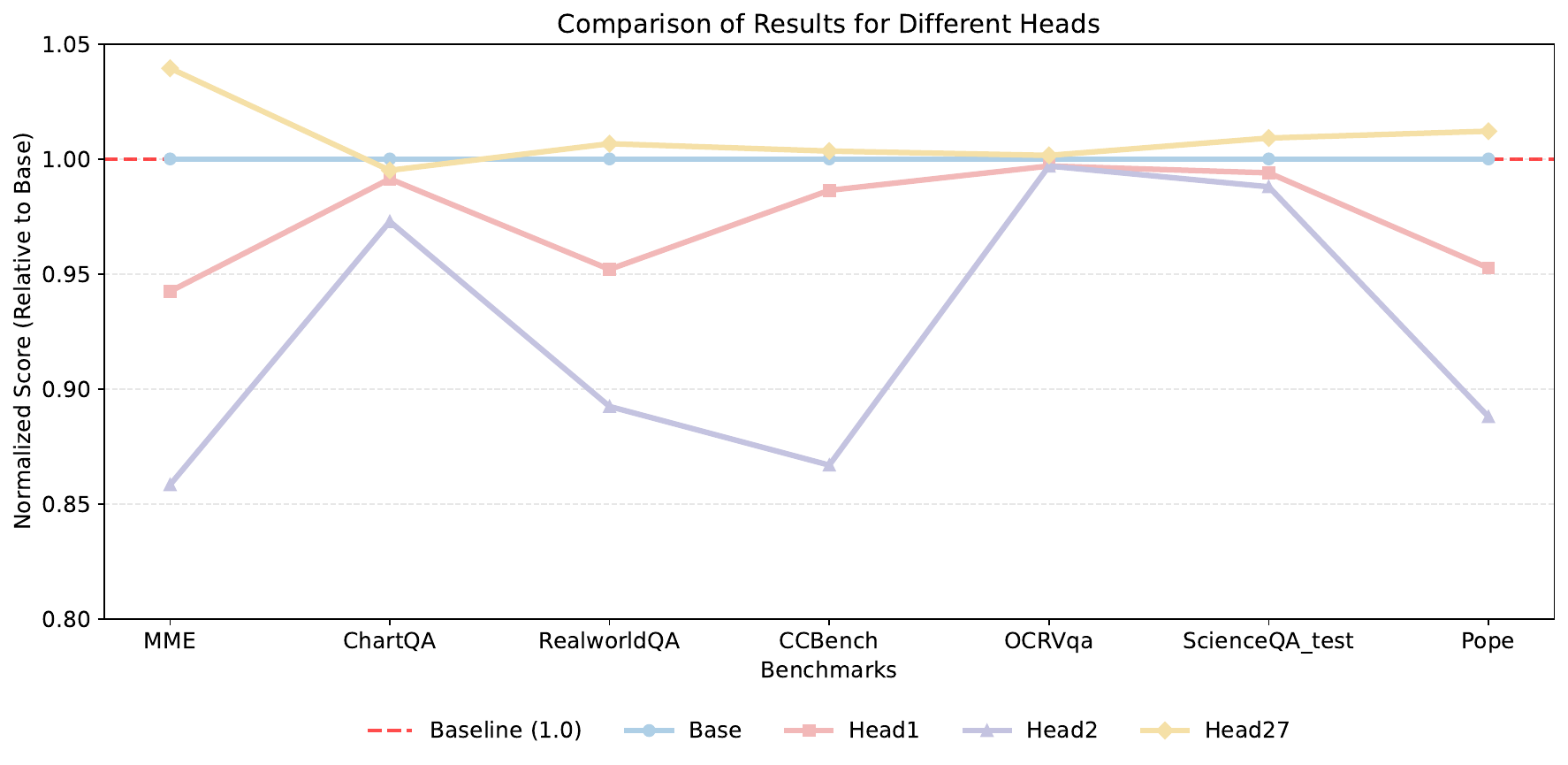}
  \caption{\textbf{Visual Head Ablation Study.} Comparison of results across different benchmarks relative to the Base model. The red dashed line indicates the baseline performance (1.0).}
  \label{fig:head_ablation}
\end{figure}
As discussed in the main text, masking the visibility of visual tokens for specific attention heads results in a consistent pattern of performance variation across diverse tasks. To further verify the generalizability  of this finding, we expanded our evaluation scope in this section. Specifically, we selected three representative attention heads (Head 1, Head 2, and Head 27) for in-depth assessment across a broader suite of benchmarks, including MME\cite{mme}, ChartQA\cite{chartqa}, RealworldQA\cite{realworldqa}, CCBench\cite{ccbench}, OCRVqa\cite{ocrvqa}, ScienceQA-IMG\cite{scienceqa}, and POPE\cite{pope}. Note that certain benchmarks were excluded from this analysis due to Out-Of-Memory (OOM) errors encountered during evaluation to ensure experimental feasibility.

\begin{figure}[ht]
    \centering
    % 子图 1
    \begin{subfigure}[b]{0.48\linewidth}
        \centering
        \includegraphics[width=\linewidth]{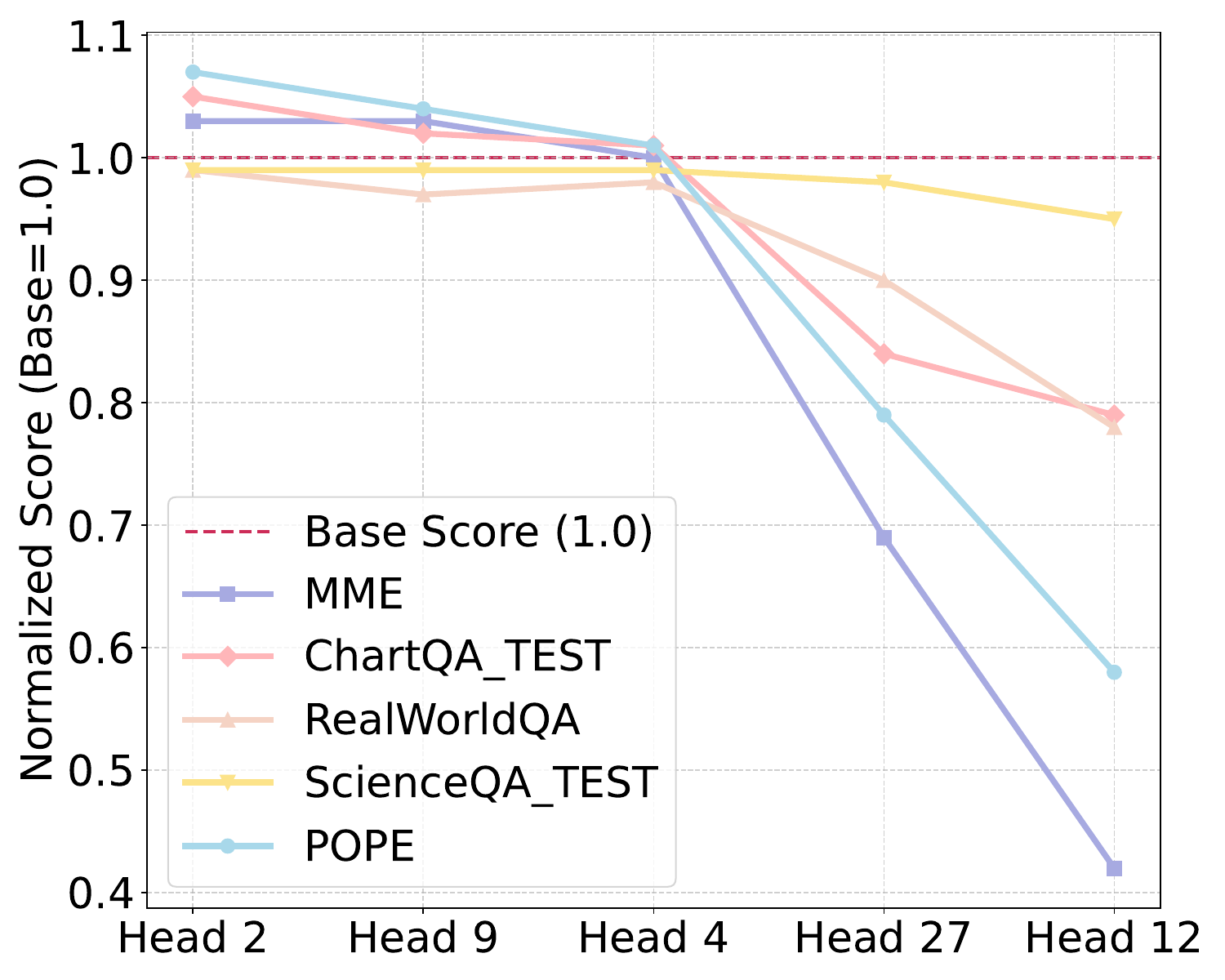}
        \caption{LLava 1.5-7b}
        \label{fig:llava}
    \end{subfigure}
    \hfill
    % 子图 2
    \begin{subfigure}[b]{0.48\linewidth}
        \centering
        \includegraphics[width=\linewidth]{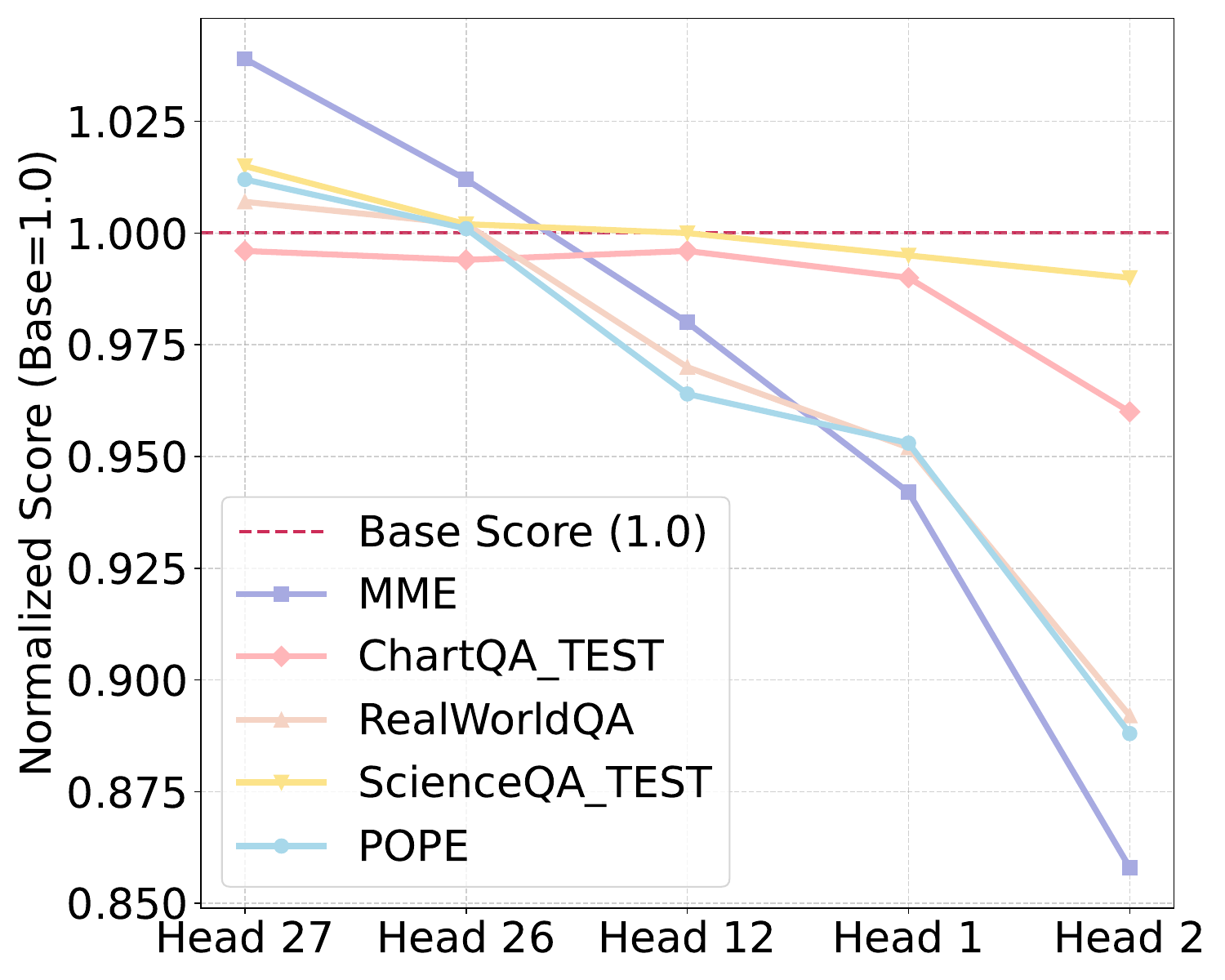}
        \caption{Qwen 2.5vl-7b}
        \label{fig:qwen}
    \end{subfigure}
    \caption{Cross-model analysis of visual head ablation.}
    \label{fig:ablation1}
\end{figure}

The results, illustrated in Figure \ref{fig:head_ablation}, strongly corroborate our conclusions in the main text regarding the functional specificity of visual heads. Specifically, Head 2 demonstrates a critical role across the majority of benchmarks; masking this head causes substantial performance degradation, particularly on tasks demanding fine-grained visual perception such as CCBench and POPE (where normalized scores drop below 0.9). In contrast, masking Head 27 yields no negative impact and even leads to marginal performance gains on specific tasks, suggesting that this head likely encodes redundant information or visual noise. Furthermore, the varying sensitivity to head masking across benchmarks, exemplified by the drastic fluctuations on CCBench versus the relative stability on ScienceQA, further underscores the high dependency of complex reasoning tasks on specific critical visual heads.
We validated head importance on \textbf{5 datasets} across \textbf{two models} (LLaVA, Qwen). As shown in \textbf{Fig.~\ref{fig:ablation1}}, head rankings remain consistent across datasets for each model. This proves our method is \textbf{dataset-independent} and relies on intrinsic model features. We will \textbf{open-source} these universal head weights for mainstream MLLMs in the future.
\vspace{-1em}

% % 
% Having the supplementary compiled together with the main paper means that:
% % 
% \begin{itemize}
% \item The supplementary can back-reference sections of the main paper, for example, we can refer to \cref{sec:intro};
% \item The main paper can forward reference sub-sections within the supplementary explicitly (e.g. referring to a particular experiment); 
% \item When submitted to arXiv, the supplementary will already included at the end of the paper.
% \end{itemize}
% % 
% To split the supplementary pages from the main paper, you can use \href{https://support.apple.com/en-ca/guide/preview/prvw11793/mac#:~:text=Delete%20a%20page%20from%20a,or%20choose%20Edit%20%3E%20Delete).}{Preview (on macOS)}, \href{https://www.adobe.com/acrobat/how-to/delete-pages-from-pdf.html#:~:text=Choose%20%E2%80%9CTools%E2%80%9D%20%3E%20%E2%80%9COrganize,or%20pages%20from%20the%20file.}{Adobe Acrobat} (on all OSs), as well as \href{https://superuser.com/questions/517986/is-it-possible-to-delete-some-pages-of-a-pdf-document}{command line tools}.
\section{More results compared with other methods}
We extended our evaluation to include comparisons with DART~\cite{dart} and VisionZip~\cite{visionzip}. As detailed in Table \ref{tab:baselines}, HAWK consistently outperforms both baselines. Notably, at a 0.8 pruning ratio, HAWK preserves \textbf{95.8\%} of the original performance, surpassing VisionZip (\textbf{89.2\%}) and DART (\textbf{80.4\%}) by substantial margins of \textbf{6.6\%} and \textbf{15.4\%}, respectively. 
\begin{table}[t]
    \centering
    \small
    \caption{Comparison with baselines at varying pruning ratios. }
    \vspace{-0.5em}
    \resizebox{\linewidth}{!}{%
    \begin{tabular}{l|c|ccccc|c}
    \toprule
    Method & Ratio & MME & ChartQA & TextVQA & AI2D & RealWorld & Avg. Rel. \\
    \midrule
    \textbf{Full Model} & 1.0 & 2315 & 86.2 & 85.2 & 80.7 & 67.7 & 100\% \\
    \midrule
    VisionZip & 0.6 & 2308 & 78.0 & 75.3 & 79.9 & 69.8 & 96.1\% \\
    DART & 0.6 & 2260 & 64.7 & 73.8 & 73.2 & 66.1 & 89.5\% \\
    \textbf{HAWK} & 0.6 & \textbf{2313} & \textbf{83.6} & \textbf{85.0} & \textbf{79.9} & 67.6 & \textbf{99.1\%} \\
    \midrule
    VisionZip & 0.8 & 2182 & 62.1 & 70.1 & 77.9 & 68.2 & 89.2\% \\
    DART & 0.8 & 2135 & 47.3 & 67.2 & 68.3 & 62.0 & 80.4\% \\
    \textbf{HAWK} & 0.8 & \textbf{2311} & \textbf{76.8} & \textbf{83.0} & \textbf{78.1} & 65.0 & \textbf{95.8\%} \\
    \midrule
    VisionZip & 0.9 & 1923 & 43.5 & 61.4 & 71.0 & 63.7 & 77.5\% \\
    DART & 0.9 & 1992 & 34.6 & 59.8 & 65.2 & 56.9 & 72.2\% \\
    \textbf{HAWK} & 0.9 & \textbf{2101} & \textbf{65.2} & \textbf{79.8} & \textbf{75.3} & 60.4 & \textbf{88.5\%} \\
    \bottomrule
    \end{tabular}%
    }
    \label{tab:baselines}
\end{table}

\section{Computation Cost of Head Weights and Description of Benchmarks}
The offline ablation is a negligible \textbf{one-time} cost. As shown in Table \ref{tab:head_overhead}, calculating head importance takes only \textbf{0.39--0.83 hours} per dataset on 8  GPUs. In contrast, training-based methods require hundreds of hours. Table\ref{tab:benchmark_overview} provides a brief overview of the benchmarks.

\begin{table}[ht]
    \centering
    \vspace{-0.5em}
    \caption{Overhead of head weight calculation on LLaVA-1.5-7B}
    \label{tab:head_overhead}
    \vspace{-1em}
    \begin{tabular}{lcc}
    \toprule
    \textbf{Dataset} & \textbf{Time (hours)} & \textbf{GPU Memory (GB)} \\
    \midrule
    MME         & 0.56 & 14.30 \\
    POPE        & 0.83 & 14.21 \\
    RealWorldQA & 0.39 & 14.33 \\
    \bottomrule
    \end{tabular}
    \vspace{-1em}
\end{table}
% \section{Brief Description of Benchmarks}
% To comprehensively evaluate the effectiveness and generalizability of our method, we select twelve representative benchmarks covering a wide spectrum of multi-modal capabilities. As summarized in Table \ref{tab:benchmark_overview}, these datasets are categorized into image-based benchmarks and video-based benchmarks. The image-based category is further divided into general visual reasoning tasks (e.g., POPE, MME) and text/diagram-oriented tasks (e.g., TextVQA, ChartQA) to assess fine-grained perception. Additionally, we include two video benchmarks (Video-MME and WorldSense) to evaluate the model's temporal understanding and reasoning abilities.
\begin{table*}[t]
  \centering
  % 调整表格宽度以适应页面
  \resizebox{\textwidth}{!}{
  \begin{tabular}{l l p{12.8cm}}
    \toprule
    \textbf{Benchmark} & \textbf{Task Type} & \textbf{Description} \\
    \midrule
    \multicolumn{3}{l}{\textit{\textbf{Image-based Benchmarks}}} \\
    \midrule
    POPE~\cite{pope} & Object Hallucination & Evaluates object hallucination ratios using random, popular, and adversarial sampling settings. \\
    HallBench~\cite{hallbench} & Visual Hallucination & Focuses on visual illusion and hallucination detection, requiring detailed image context reasoning. \\
    MME~\cite{mme} & Comprehensive Evaluation & A comprehensive suite covering 14 subtasks, including perception and cognition capabilities. \\
    ScienceQA-IMG~\cite{scienceqa} & Science VQA & Contains multimodal science questions with annotated reasoning explanations and image contexts. \\
    RealWorldQA~\cite{realworldqa} & Spatial Reasoning & Evaluates spatial perception and reasoning capabilities in diverse real-world environments. \\
    CCBench~\cite{ccbench} & Cultural Understanding & A benchmark designed to assess the model's understanding of Chinese cultural contexts and general knowledge. \\
    \midrule
    TextVQA~\cite{textvqa} & OCR-VQA & Requires reading and reasoning about text embedded in natural images to answer questions. \\
    OCRVQA~\cite{ocrvqa} & OCR-VQA & Focuses on visual question answering based on text-rich images such as book covers. \\
    ChartQA~\cite{chartqa} & Chart Understanding & Involves reasoning over complex charts and graphical data, requiring numerical and logical analysis. \\
    AI2D~\cite{ai2d} & Diagram Understanding & Evaluates the comprehension of science diagrams and textbook illustrations. \\
    \midrule
    \multicolumn{3}{l}{\textit{\textbf{Video-based Benchmarks}}} \\
    \midrule
    Video-MME~\cite{videomme} & Video Understanding & A comprehensive benchmark for long-duration video understanding across diverse domains. \\
    WorldSense~\cite{worldsense} & Video Reasoning & Assesses the understanding of physical laws, world dynamics, and causal reasoning in videos. \\
    \bottomrule
  \end{tabular}
  }
  \caption{\textbf{Overview of Evaluation Benchmarks.} We select ten image-based benchmarks (divided into general/reasoning and text/diagram categories) and two video-based benchmarks to comprehensively evaluate the model's capabilities.}
  \label{tab:benchmark_overview}
\end{table*}

\section{Additional Visualization Results}
\begin{figure*}[h]
    \centering
    \captionsetup{type=figure} % 确保 caption 编号正确
    
    % ============================================================
    % Example 1: ChartQA (Hero Example)
    % 第一页只放这一个，图片可以给得非常大
    % ============================================================
    \begin{minipage}{\textwidth}
        \centering
        % 图片：设置为页面宽度的 49%，高度最大限制放宽到 10cm，保证细节清晰
        \includegraphics[width=0.49\linewidth, height=8cm, keepaspectratio]{pict/chartqa.png}
        \hfill
        \includegraphics[width=0.49\linewidth, height=8cm, keepaspectratio]{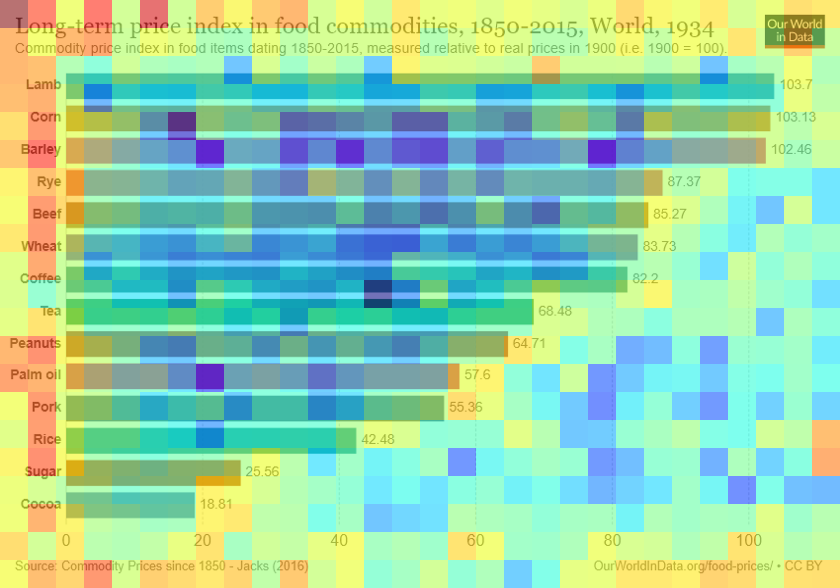}
        
        \vspace{1.5mm}
        
        % 文本框：因为只有这一个例子，文字可以不用缩写，完整展示
        \begin{tcolorbox}[colback=gray!5, colframe=gray!60, boxrule=0.5pt, arc=2mm, title=\textbf{Example 1: ChartQA (Chart Understanding )}]
            \small % 字体可以舒适的大小
            \textbf{Query:} \textit{How many food item is shown in the bar graph?}
            \par\noindent\rule{\linewidth}{0.1pt}
            
            \textbf{Divprune:} The bar graph shows \textcolor{cred}{seven} different food items: Lamb, Barley, Beef, Tea, Pork oil, Sugar, and Cocoa. \\
            \textbf{CDPruner:} The bar graph shows \textcolor{cred}{six} different food items. The items listed are:
            1. Palm oil, 2. Pork, 3. Rice, 4. Sugar, 5. Cocoa, 6. Wheat. \\
            \textbf{HAWK (Ours):} The bar graph shows \textcolor{cgreen}{\textbf{14}} different food items. These are:
            Lamb, Corn, Barley, Rye, Beef, Wheat, Coffee, Tea,Peanuts, Palm oil, Pork, Rice, Sugar, Cocoa. 
            So, there are 14 food items shown in the bar graph.
        \end{tcolorbox}
    \end{minipage}

    % 主 Caption
\caption{\textbf{Qualitative Results: Heatmaps and Response Comparison.} 
We present attention heatmaps to visualize the visual tokens retained by our HAWK, \textbf{where redder regions indicate higher attention scores}, alongside a qualitative comparison of the generated responses against other baseline methods. 
(Figure continued on the next page...)}
    \label{fig:supp_qualitative_results}
\end{figure*}

\begin{figure*}[t]
    \ContinuedFloat % <--- 关键：继承上一个图的编号 (例如 Figure 8 -> Figure 8 continued)
    \centering
    
    % ============================================================
    % Example 2
    % ============================================================
    \begin{minipage}{\textwidth}
        \centering
        % 图片高度稍微收敛一点 (8cm)，以便一页能舒服地放下两个
        \includegraphics[width=0.49\linewidth, height=8cm, keepaspectratio]{pict/mme.png}
        \hfill
        \includegraphics[width=0.49\linewidth, height=8cm, keepaspectratio]{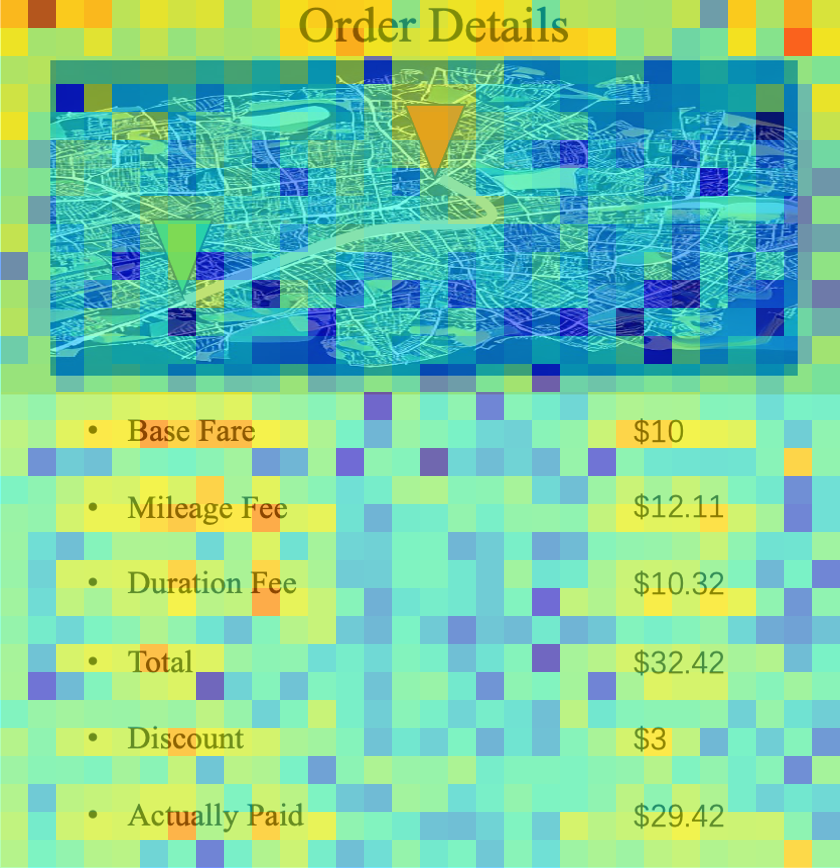}

        \begin{tcolorbox}[colback=gray!5, colframe=gray!60, boxrule=0.5pt, arc=2mm, title=\textbf{Example 2: [MME/ Comprehensive Evaluation]}]
            \small
                        \textbf{Query:} \textit{Here are the order details for my taxi ride. Should I actually pay \$29.42? Please answer yes or no.}
            \par\noindent\rule{\linewidth}{0.1pt}
            \textbf{Divprune:} \textcolor{cred}{No.} \\
            \textbf{CDPruner:} \textcolor{cred}{No.} \\
            \textbf{HAWK (Ours):} \textcolor{cgreen}{\textbf{Yes.}}
        \end{tcolorbox}
    \end{minipage}

    % ============================================================
    % Example 3
    % ============================================================
    \begin{minipage}{\textwidth}
        \centering
        \includegraphics[width=0.49\linewidth, height=8cm, keepaspectratio]{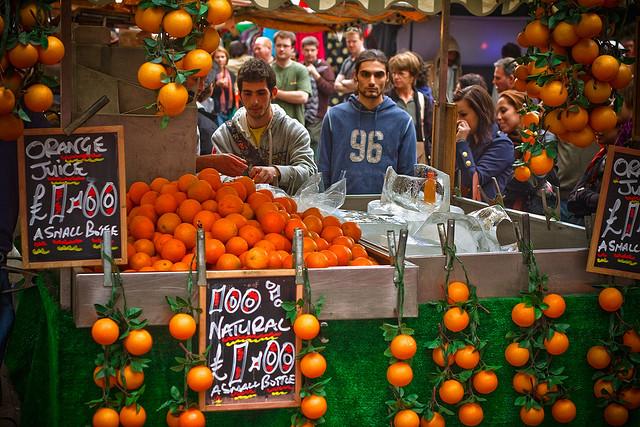}
        \hfill
        \includegraphics[width=0.49\linewidth, height=8cm, keepaspectratio]{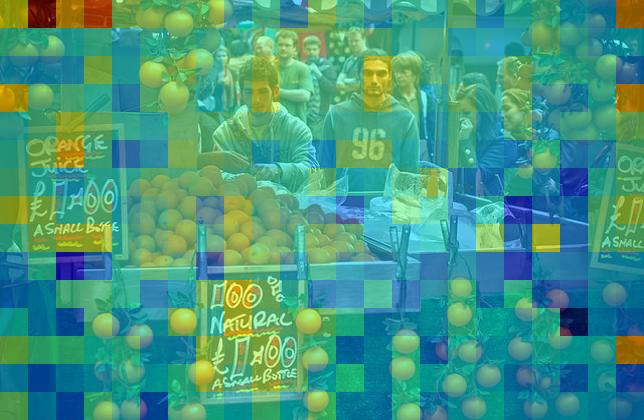}
        
        \begin{tcolorbox}[colback=gray!5, colframe=gray!60, boxrule=0.5pt, arc=2mm, title=\textbf{Example 3: [POPE/ Object Hallucination]}]
            \small
                        \textbf{Query:} \textit{Is there a bottle in the image?}
            \par\noindent\rule{\linewidth}{0.1pt}
            \textbf{Divprune:} \textcolor{cgreen}{Yes},there is a bottle in the image. It appears to be a glass bottle with a label, located on the right side of the image, near the top. \\
            \textbf{CDPruner:} \textcolor{cred}{No},there is no bottle in the image. The image shows a display of oranges and a sign advertising "Orange Juice" for sale. The sign mentions "A Small Bottle," but there are no actual bottles visible in the image. \\
            \textbf{HAWK (Ours):} \textcolor{cgreen}{\textbf{Yes}},there is a bottle in the image. It is located on the right side of the image, near the top. The bottle appears to be part of the display for the orange juice stand.
        \end{tcolorbox}
    \end{minipage}

    % 这里的 caption 可以为空，或者写简单的说明
    \caption[]{Continued. }
\end{figure*}


\end{document}